\documentclass[runningheads]{llncs}
\usepackage{eccv}

\usepackage{eccvabbrv}
\usepackage{graphicx}
\usepackage{booktabs}
\usepackage[accsupp]{axessibility}  

\usepackage{hyperref}
\usepackage{orcidlink}

\usepackage{amssymb}
\newcommand{\cmark}{\ding{51}}
\newcommand{\xmark}{\ding{55}}
\usepackage{pifont}
\usepackage{placeins} 
\usepackage{eccvabbrv}
\usepackage{graphicx}
\usepackage{booktabs}
\usepackage[accsupp]{axessibility}
\usepackage{algorithm}
\usepackage{algpseudocode}
\usepackage{textcomp}
\usepackage{gensymb}
\usepackage{etoc}
\etocchecksemptiness
\makeatletter
\def\authcount#1{}
\etocsetlevel{title}{-1}
\etocsetlevel{author}{-1}
\makeatother
\usepackage{pdfpages}

\newif\ifincludesupp
\includesupptrue   

\begin{document}

\title{Filterless Snapshot Hyperspectral Imaging using Guided Patch Diffusion} 

\author{Dean Hazineh\orcidlink{0000-0002-8762-1621} \and
    Luca Sacchi \and
    Davide Cassara\orcidlink{0009-0004-4731-6608} \and \\ 
    Federico Capasso\orcidlink{0000-0003-4534-8249} \and
    Todd Zickler\orcidlink{0000-0002-3853-1558}
}

\authorrunning{D.~Hazineh et al.}

\institute{
    Harvard University\\
    \email{dhazineh@g.harvard.edu}
}

\maketitle
\begin{abstract}
    We consider the problem of reconstructing a $H\times W\times 31$ hyperspectral image from a $H\times W$ grayscale snapshot measurement that is captured using only a single diffractive lens and a filterless panchromatic photosensor. This problem is severely ill-posed, but we present a model that produces high-quality results in simulation and experiment. We make efficient use of limited training data by creating a conditional denoising diffusion model that operates on small patches in a shift-invariant manner. During inference, we synchronize per-patch hyperspectral predictions using guidance by physical consistency with the system's optical point spread function. Our experiments reveal that the patch size can be as small as the point spread function, with local optical cues being the main source of information about complete spectra. Also, by drawing multiple samples, our model provides per-pixel uncertainty estimates that strongly correlate with reconstruction error.
\end{abstract}

\begin{figure}[t]
  \centering
  \begin{subfigure}[t]{0.445\linewidth}
    \centering
    \includegraphics[width=\linewidth]{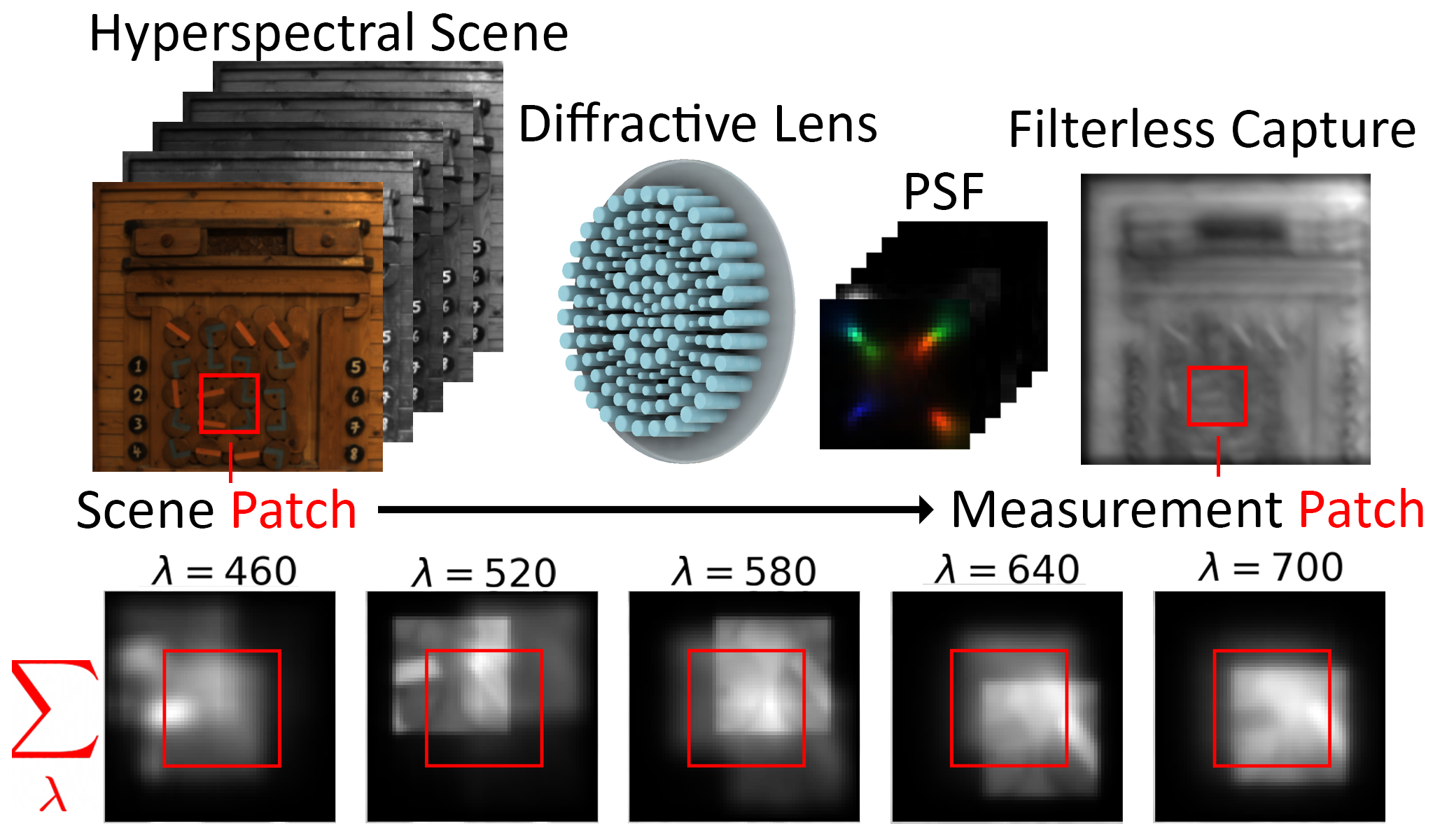}
    \caption{A hyperspectral scene is imaged through a diffractive lens onto a filterless photosensor. Bottom: the PSF blurs and shifts each wavelength differently, and the grayscale measurement is their sum. The patch boundary is denoted by the red square.}
    \label{fig:overview}
  \end{subfigure}
  \hfill
  \begin{subfigure}[t]{0.545\linewidth}
    \centering
    \includegraphics[width=\linewidth]{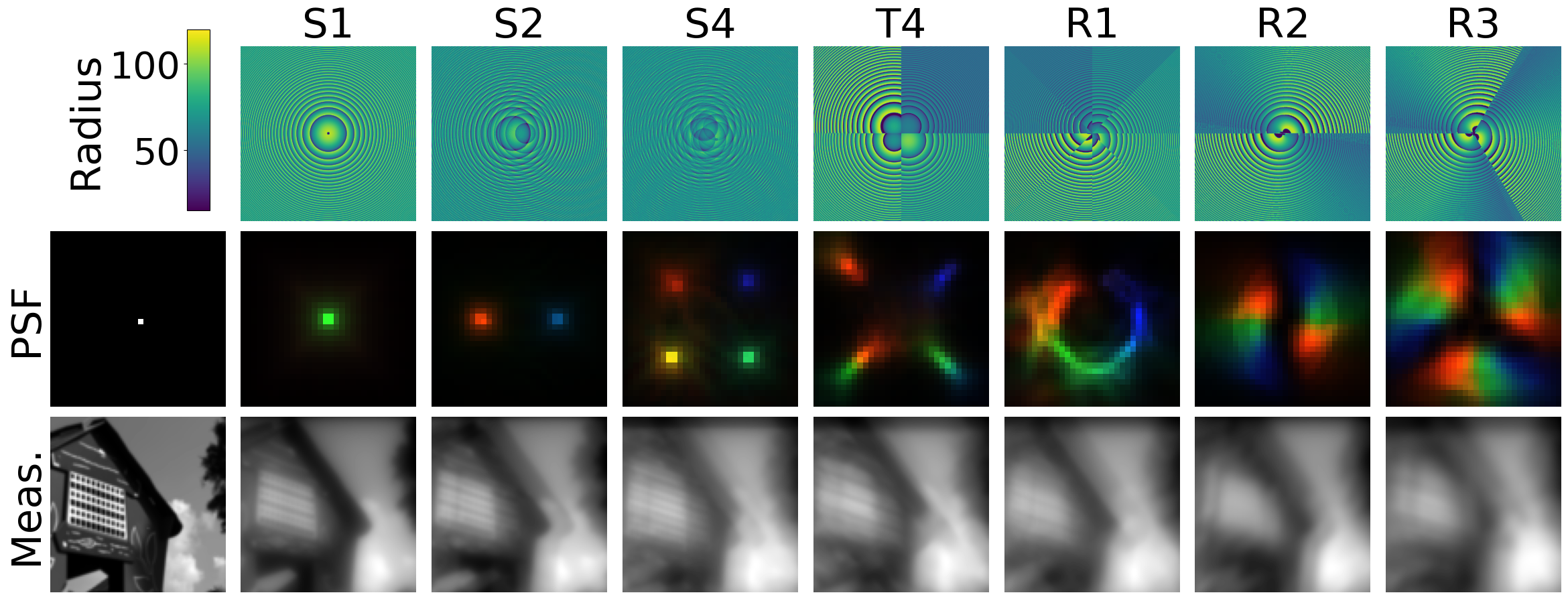}
    \caption{Flat optic designs used in our experiments. Top: nanocylinder radius patterns (nm). Middle: PSFs (projected to RGB). Bottom: example measurements. The leftmost column shows the ideal all-in-focus measurement for reference.}
    \label{fig:metalens}
  \end{subfigure}
  \caption{Forward optical model and flat optic designs.}
\end{figure}

\section{Introduction}
\label{sec:intro}
Snapshot hyperspectral cameras capture detailed spectral information about a scene at a single moment in time. They offer a richer representation than standard RGB images and are useful for scientific detection and classification. Generally, these cameras have two coupled components: an optical assembly that encodes spatial and spectral information onto a photosensor, and a digital decoder that reconstructs the hyperspectral image (HSI) from the resulting measurement. To better condition the reconstruction problem, existing snapshot designs typically use one or more of the following strategies~\cite{DingWang2024}: complex, multi-stage optics; color filter arrays on the photosensor; and/or photosensors with more pixels than the intended HSI spatial resolution.

In this paper, we explore a minimalist snapshot scenario that is less well-posed and largely underexplored. Our goal is to reconstruct a $H\times W\times 31$ HSI using only: (\emph{i}) a filterless panchromatic photosensor with $H \times W$ pixels, the same number of measurement pixels as output pixels; and (\emph{ii}) a single flat optic lens, such as a diffractive optical element or a metalens. This scenario is interesting because it could enable a new class of snapshot hyperspectral cameras with distinct trade-offs between optical and computational complexity. Using a single flat optic improves compactness, and removing spectral filter mosaics on the sensor increases light throughput and avoids interleaved sub-sampling. 

We can design the flat optic to induce purposeful chromatic aberration that mixes spatial and spectral information into the filterless measurement, as shown in the top of \cref{fig:overview}. However, reconstructing the HSI from this mixture is severely ill-posed. It requires an effective prior from a deep learning model, but this is hard to train because, compared to RGB images, ground-truth hyperspectral images are relatively scarce. Patch-based generative diffusion models have recently emerged as a promising solution for learning strong priors from small datasets~\cite{wang2023, hu2024a, hu2024b, ozdenizci2022,han2024multistable}, but conventional patch-based processing is particularly difficult to apply here. As shown in the bottom of \cref{fig:overview}, the measurement is formed by convolution with an optical point-spread function (PSF) with a spatial extent that is comparable to the patch size. This means that a significant amount of the target hyperspectral signal is scattered outside of its corresponding measurement patch (red square), making per-patch reconstruction very ambiguous. As far as we know, no prior patch-based approach to reconstruction, diffusion-based or otherwise, has been shown to have success in these conditions.

We address this challenge and introduce a patch diffusion model that succeeds at our task. We overcome the ambiguity associated with patching by adopting global diffusion guidance during inference, where patches are iteratively denoised in parallel and then assembled into full-sized HSIs that are forced to be optically consistent with the full-size measurement. We find that this resolves patch-based ambiguities and provides better results than any previous model applied to our task. Like any patch-based approach, our model has the advantage of being able to operate on any image size once it is trained, and we also find that it provides useful per-pixel uncertainty estimates for its spectral predictions that strongly correlate with reconstruction error. We extensively evaluate our method in simulation and provide an experimental prototype as a proof-of-concept. 

\section{Related Work}
\label{sec:related}
Training diffusion models on image patches substantially reduces data requirements. Unconditional patch models can generate high-quality RGB images from as few as $5$K samples~\cite{wang2023}, and patch-based diffusion has been applied to inverse problems including deblurring~\cite{hu2024a} and deweathering~\cite{ozdenizci2022} using comparably small datasets. However, these works consider spatially compact degradations. We show that patch-based diffusion can handle very large degradation kernels by leveraging inference-time guidance.

In aerial remote sensing, where large hyperspectral datasets exist, unconditional diffusion models have been trained from scratch to learn representations for classification~\cite{Chen2023, sigger2024}. For natural scenes, datasets are far smaller---ARAD1K~\cite{Arad_2022_CVPR_recovery}, for example, contains only $950$ HSIs---and existing methods instead rely on frozen, pre-trained RGB diffusion models for plug-and-play HSI restoration~\cite{Pang2024, Zeng2024} or compressed sensing~\cite{Pan2024}. In contrast, we train conditional hyperspectral diffusion models from scratch on natural scene patches, learning spatial-spectral patch priors without overfitting to the small available datasets. To our knowledge, similar models have not been explored in this context.

Recovering a HSI from a single unfiltered measurement has traditionally required complex optical assemblies. CASSI~\cite{Wagadarikar2008, Meng2020}, for example, uses coded masks, dispersive elements, and oversized photosensors (\eg, $H\times(W{+}30)$ measurement pixels for an $H{\times}W{\times}31$ output). Other approaches rely on spectral filters such as RGB Bayer patterns~\cite{Arad2016, Yuanhao2022, Arad_2022_CVPR_recovery, Zhang2023b} (sometimes paired with diffractive lenses~\cite{Jeon2019, zhang2023, Lv_2023_ICCV}) or optimized color filter arrays~\cite{Nie2018, Monakhova20, Salesin2022, Li2023}. However, spectral filters discard a large fraction of incoming light---a Bayer mosaic, for example, captures only one of three color channels per pixel. Our approach requires neither multi-component optics nor spectral filters: we place a single flat diffractive lens in front of a bare panchromatic photosensor and rely on a strong diffusion-based decoder to recover the HSI. Concurrently, Liu~\etal~\cite{Liu2025} explore a similar, filterless reconstruction problem but using a SLM-based optical encoder, providing further evidence that this problem class is tractable.

Additionally, we leverage our diffusion model to produce uncertainty estimates for predicted spectra. Concurrent work from Romero~\etal~\cite{romero2025} also explores uncertainty quantification in HSI reconstructions using diffusion posterior sampling, though applied to RGB inputs.

\section{Methods}
Let $\mathbf{x} \in \mathbb{R}^{H \times W \times C}_{\geq 0}$ denote a hyperspectral image (HSI), representing the far-field scene’s undistorted spatial–spectral radiance mapped to the photosensor plane by an ideal lens focused at infinity.
This representation accounts for geometric magnification and spatial discretization to the sensor’s pixel size. We define the associated measurement $\mathbf{y}\in\mathbb{R}^{H\!\times\! W}_{\geq0}$ that is induced by a diffractive lens using the element's wavelength-dependent point-spread function (PSF) $f(u,v,\lambda)$ as
\begin{equation}
    \mathbf{y}(u,v)\! =\! \mathcal{M}(\mathbf{x}) \!=\! \sum_\lambda o(\lambda) \cdot f(u,v,\lambda) \!\underset{(u,v)}{*}\! \mathbf{x}(u,v,\lambda),
    \label{eq:measurement}
\end{equation}
where $*$ denotes 2D convolution over the spatial dimensions and $o(\lambda)$ corresponds to the spectral response of the photosensor. Throughout, we assume that the PSF is shift-invariant. A measurement is thus a non-invertible linear optical encoding of a 3D hyperspectral cube to a 2D image. Our objective is to model the conditional distribution $p(\mathbf{x}\mid\mathbf{y})$ and to sample hyperspectral images $\mathbf{x}$ that are consistent with the observed measurement $\mathbf{y}$. We do this by training a patch-based diffusion model and introducing a guided sampling algorithm that synchronizes patch predictions into measurement-consistent full-field HSIs.

\subsection{Optical Encoding}
\label{ssec:metalens} 
In our experiments, we test the eight point-spread functions (PSFs) shown in the middle row of \cref{fig:metalens} to understand what type of optical encoder is most effective for our task. These PSFs vary in the extent to which they spread spectral information across space, producing differently-blurred measurements. More concentrated PSFs (left) produce sharper images that preserve spatial detail but encode spectral information less effectively than more dispersive PSFs (right). Because reconstruction requires both high spatial and spectral accuracy, it is not obvious which type of PSF will perform the best in our filterless scenario. 

All of these PSFs can be physically realized using a diffractive lens known as a \emph{metalens}---a transparent glass sheet patterned with nanoscale cylinders of equal height and varying widths \cite{khorasaninejad2016b,Khorasaninejad2017r}. The radius of each nanocylinder controls the local phase-delay, and each of the PSFs results from a different arrangement of radii (top). We design a subset of the lenses, labeled with prefix ``S'',  using spatial multiplexing to produce a quasi-stationary, multi-foci effect \cite{arbabi2016}. The other lenses, labeled with prefix ``T'' or ``R'', are designed using angular multiplexing to produce PSFs whose focal spots shift or rotate as a function of wavelength. The designs for ``R2'' and ``R3'' have been used previously for RGB-to-hyperspectral imaging and follow from \cite{Jeon2019}. The PSF for each lens is computed using a wave-optics simulator~\cite{DFlat, Hazineh2023, Brookshire24}. See Supp.~Sec.~S1 for more details. 

\begin{figure}[t]
    \centering
    \includegraphics[width=\textwidth]{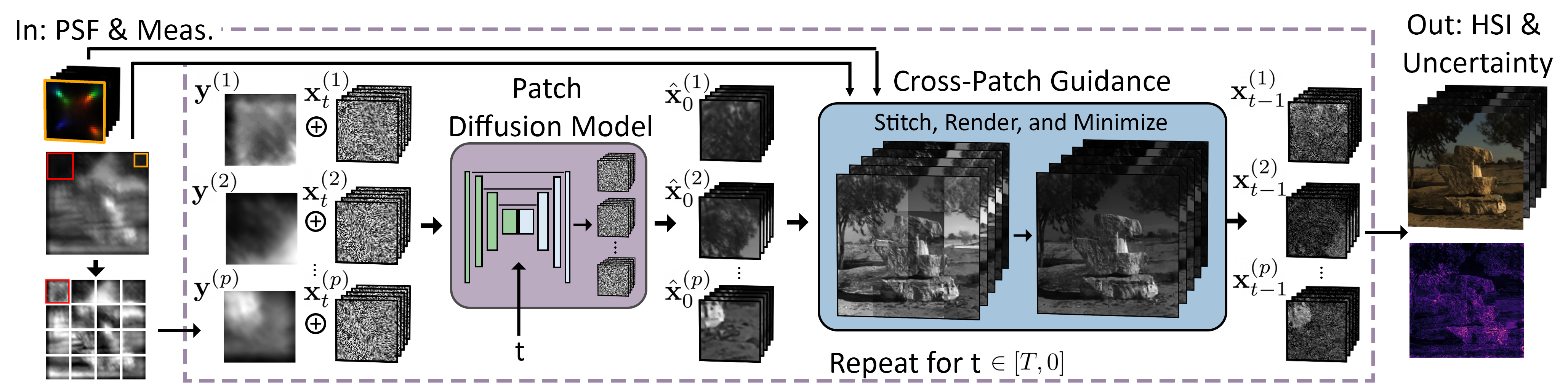}
    \caption{
    Overview of the guided sampling pipeline. The input measurement is split into patches, each concatenated with a noise sample and denoised. Intermediate patch predictions are stitched into a full-field HSI and passed through the forward model to compute a guidance gradient, which updates all patches before the next reverse diffusion step. Multiple runs with different random seeds yield per-pixel uncertainty.}
    \label{fig:restoration_algorithm}
\end{figure}

\subsection{Denoising Diffusion on Patches}
\label{ssec:Diffusion}
We use a conditional denoising diffusion model~\cite{Ho2020} to sample from $p(\mathbf{x}|\mathbf{y})$. The forward process corrupts a clean HSI $\mathbf{x}_0$ with Gaussian noise $\boldsymbol{\epsilon} \sim \mathcal{N}(\mathbf{0}, \mathbb{I})$ according to $\mathbf{x}_t = \sqrt{\alpha_t}\mathbf{x}_0 + \sqrt{1-\alpha_t}\boldsymbol{\epsilon}$, where $\alpha_t$ is a variance schedule. A network $\boldsymbol{\epsilon}_\theta(\mathbf{x}_t, t; \mathbf{y})$ is trained to predict $\boldsymbol{\epsilon}$ from the noisy sample $\mathbf{x}_t$ at timestep $t$, conditioned on the measurement $\mathbf{y}$, by minimizing $\mathbb{E}_{\mathbf{x}_0, \boldsymbol{\epsilon}, t}\!\left[\|\boldsymbol{\epsilon} - \boldsymbol{\epsilon}_\theta(\mathbf{x}_t, t; \mathbf{y})\|^2\right]$. Samples are drawn via the reverse process~\cite{song2022}:\begin{equation}
    \begin{aligned}
        \hat{\mathbf{x}}_0(\mathbf{x}_t) &= \frac{\mathbf{x}_t - \sqrt{1-\alpha_t}\, \boldsymbol{\epsilon}_\theta}{\sqrt{\alpha_t}},\\
        \mathbf{x}_{t-1} &= \sqrt{\alpha_{t-1}}\, \hat{\mathbf{x}}_0(\mathbf{x}_t) + \sqrt{1 - \alpha_{t-1} - \sigma_t^2}\, \boldsymbol{\epsilon}_\theta + \mathbf{w},
    \end{aligned}
    \label{eq:reverse}
\end{equation}
where $\mathbf{w} \sim \mathcal{N}(\mathbf{0}, \sigma_t^2\mathbb{I})$ and $\sigma_t$ controls stochasticity. The intermediate estimate $\hat{\mathbf{x}}_0(\mathbf{x}_t)$ plays a central role in our guided sampling procedure (\cref{ssec:Guidance}).

Rather than denoising full HSIs at once, we model the conditional distribution at the patch level. Because the PSF support is comparable to the patch size, each measurement patch contains signal from neighboring HSI regions while also losing signal scattered to adjacent patches (\cref{fig:overview}). Despite this crosstalk, we find that a patch-based diffusion model can learn a prior that compensates for the ambiguous information at patch boundaries. For training data, we use captured HSIs from a ground truth hyperspectral camera and pre-render the corresponding measurements via \cref{eq:measurement}. We then train our models using pairs of patches $(\mathbf{x}_0^{(i)}, \mathbf{y}^{(i)})$ randomly cropped from the full-field HSI–measurement pairs. We implement conditioning through concatenation, as shown in \cref{fig:restoration_algorithm}. Each patch pair is max-normalized prior to training (see Supp.~Sec.~S3.2); the resulting per-patch scale ambiguity is resolved during guided sampling as described next. 

\subsection{Sampling with PSF Guidance}
\label{ssec:Guidance}
Applying the denoising formulation in \cref{eq:reverse} to patches produces hyperspectral patch predictions $\hat{\mathbf{x}}_0^{(i)}$ at intermediate time steps $t$. We use these predictions to guide the denoising step and impose additional constraints when sampling $\mathbf{x}_{t-1}^{(i)}$ from $\mathbf{x}_t^{(i)}$~\cite{chung2024, chung2024b}. In particular, we enforce that all intermediate hyperspectral patches stitch together into a full-field HSI that is optically consistent with the full-field measurement. Pseudo-code is provided in Supp.~Sec.~S3.3; here we summarize the key steps. Throughout, we use superscript $p$ to denote a $p$-element collection of patches, \eg $\mathbf{x}_t^p = \{\mathbf{x}_t^{(i)}\}_{i=1}^p$ and define a $\text{Stitch}(\cdot)$ operator that combines those patch estimates into a single full-field HSI. The operator $\mathcal{M}(\cdot)$ refers to the measurement operation in \cref{eq:measurement}.

During deployment, we split the full-field measurement $\mathbf{y}$ into non-overlapping patches $\mathbf{y}^p$, each concatenated with a per-patch noise sample $\mathbf{x}_T^p$. We then process these patches in parallel through a denoising step to obtain the intermediate denoised estimates $\hat{\mathbf{x}}_0^p$. Next, we stitch those estimates into a full-field HSI and pass it through the measurement operator. Because each patch is max-normalized during training, the denoised estimates are accurate up to an unknown per-patch scale. We recover these scales by solving the least-squares problem,
\begin{equation}
    c^p_{\text{lsq}} = \mathop{\mathrm{argmin}}_{c^p} \| \mathcal{M}( \text{Stitch}(c^p \cdot \hat{\mathbf{x}}^p_0(\mathbf{x}_t^p))) - \mathbf{y} \|^2,
    \label{eq:clsq}
\end{equation}
which admits a closed-form solution and is computed in a single pass. With the scale factors fixed at $c_{\text{lsq}}^p$, we define a guidance loss over the patch states to measure consistency with the full-field measurement,
\begin{equation}
    \mathcal{L}(\mathbf{x}_t^p, \mathbf{y}) = \| \mathcal{M}(\text{Stitch}(c^p_{\text{lsq}} \cdot \hat{\mathbf{x}}_0^p(\mathbf{x}_t^p))) - \mathbf{y} \|^2.
    \label{eq:guidance_loss}
\end{equation}
This loss provides a gradient-based correction to all patch states, applied via the modified denoising transition,
\begin{align}
    \tilde{\mathbf{x}}_t^p &= \mathbf{x}_t^p - \eta \nabla_{\mathbf{x}_t^p} \mathcal{L}\left( \mathbf{x}_t^p, \mathbf{y} \right) / \left\|\nabla_{\mathbf{x}_t^p} \mathcal{L}\left( \mathbf{x}_t^p, \mathbf{y} \right)\right\| \label{eq:guided1}\\
    \mathbf{x}_{t-1}^p &= \sqrt{\alpha_{t-1}} \hat{\mathbf{x}}_0^p\left( \tilde{\mathbf{x}}_t^p \right) + \sqrt{1 - \alpha_{t-1} - \sigma_t^2} \, \epsilon_\theta^p + w^p \label{eq:guided2},
\end{align}
where $\tilde{\mathbf{x}}_t^p$ and $\mathbf{x}_{t-1}^p$ are the updated patch states. We repeat the gradient step in \cref{eq:guided1} multiple times before the denoising step in \cref{eq:guided2}. \cref{fig:restoration_algorithm} provides an overview of the entire sampling pipeline.

Finally, by varying the initial noise samples $\mathbf{x}_T^p$, we obtain multiple plausible HSIs from the same measurement. We quantify spectral uncertainty by computing a per-pixel variance map from $N$ draws,
\begin{equation}
    \text{Uncertainty} = \sum_\lambda \mathrm{Var}\left( \{\mathbf{x}_0\}_{i=1}^N \right).
    \label{eq:uncertainty}
\end{equation}

\section{Simulation Results}
\label{sec:experiment}
    We first evaluate our reconstruction algorithm in simulation. We use the ARAD1K dataset~\cite{Arad_2022_CVPR_recovery} with a standard 900/50 train/test split, reconstructing 31 spectral channels uniformly spanning 400--700\,nm. Unless otherwise stated, we resize HSIs to $256 \times 256$ pixels, matching the resolution used by competing approaches; these methods process full-field measurements through a single network and cannot easily scale higher. Our patch-based model does not share this limitation. Grayscale measurements are rendered using the R1 PSF, which yielded the best empirical performance (\cref{ssec:lens_comparison}). We train on $64\times64$ patches and sample using 20 DDIM steps and 20 guidance iterations (\cref{ssec:model_ablations}). We report SSIM, PSNR, and SAM~\cite{yuhas1992sam} averaged over full-field reconstructions. Additional architecture, training, and inference details are provided in Supp.~Sec.~S3.

\subsection{Comparison to Other Models} 
\label{ssec:model_comparison} 
    \begin{table*}[t]
    \centering
    \scriptsize
    \setlength{\tabcolsep}{3pt}
    \begin{minipage}[t]{0.35\textwidth}
        \centering
        \captionof{table}{Filterless reconstruction on ARAD1K using the R1 PSF. Metrics: (S)AM, (SS)IM, (P)SNR.}
        \label{tab:ARAD_Table}
        \begin{tabular}{lccc}
        \toprule
        \cmidrule(lr){2-4}
        Model  & S $\downarrow$ & SS $\uparrow$ & P $\uparrow$\\
        \midrule    
        Ours  & \textbf{0.11} & \textbf{0.94} & \textbf{34.6} \\
        Ours (no guid.) & \underline{0.14} & \underline{0.92} & \underline{32.32}\\
        UNet-Regress. & 0.15 & 0.83 & 29.12 \\
        SST~\cite{SST} & 0.15 & 0.90 & 31.8\\
        SPECAT~\cite{SPECAT}  & 0.18 & 0.84 & 29.6  \\
        MST~\cite{Yuanhao2022} & 0.17 & 0.87 & 29.8 \\
        In2Set~\cite{Wang_2024_CVPR} & 0.18 & 0.86 & 30.1 \\
        DAUHST~\cite{cai2022b} & 0.17 & 0.86 & 29.7 \\
        DGSMP~\cite{huang2021}& 0.16 & 0.88 & 30.0  \\
        HDNet~\cite{hu2022}& 0.17 & 0.86 & 29.3 \\
        TSANet~\cite{Meng2020}& 0.20 & 0.87 & 29.2\\
        \bottomrule
        \end{tabular}
    \end{minipage}
    \hspace{0.02\textwidth}
    \begin{minipage}[t]{0.32\textwidth}
        \centering
        \captionof{table}{Model ablations.\\ $^\dagger$uses overlapping patches with a stride of 32 pixels}
        \label{tab:ablation}
        \begin{tabular}{@{}c c c c@{}}
        \toprule
        Patch & Resc/Guid & SS $\uparrow$ & P  $\uparrow$ \\
        \midrule
        64 & \cmark / \cmark & 0.94 & 34.7 \\
        64 & \cmark / \xmark & 0.92 & 32.2 \\
        64$^\dagger$ & \cmark / \cmark & 0.95 & 34.8 \\
        64$^\dagger$ & \cmark / \xmark & 0.93 & 33.0 \\
        \hline
        32 & \cmark / \cmark & 0.93 & 33.3 \\
        32 & \cmark / \xmark & 0.87 & 29.3 \\
        \hline
        64 & \xmark / \cmark & 0.92 & 31.8 \\
        64 & \xmark / \xmark & 0.86 & 27.5 \\
        \bottomrule
        \end{tabular}
    \end{minipage}
    \hspace{0.005\textwidth}
    \begin{minipage}[t]{0.265\textwidth}
        \centering
        \captionof{table}{Lens comparison.\\ $^*$designs from~\cite{Jeon2019}\\ }
        \label{tab:lenscomp}
        \begin{tabular}{lccc}
        \toprule
        Lens & S $\downarrow$ & SS $\uparrow$ & P $\uparrow$\\
        \midrule
        AIF & 0.20 & 0.93 & 29.8 \\
        S1 & 0.17 & 0.93 & 31.1 \\
        S2 & 0.15 & 0.94 & 33.1 \\
        S4 & 0.13 & 0.94 & 34.4 \\
        \underline{T4} & \underline{0.11} & \underline{0.94} & \underline{34.6} \\
        \textbf{R1} & \textbf{0.11} & \textbf{0.95} & \textbf{35.0} \\
        R2$^*$ & 0.11 & 0.93 & 34.0 \\
        R3$^*$ & 0.12 & 0.93 & 34.1 \\
        \bottomrule
        \end{tabular}
        \end{minipage}
    \end{table*}
    
    \begin{figure}[t]
        \centering
        \begin{subfigure}[t]{0.495\linewidth}
            \centering
            \includegraphics[width=\linewidth]{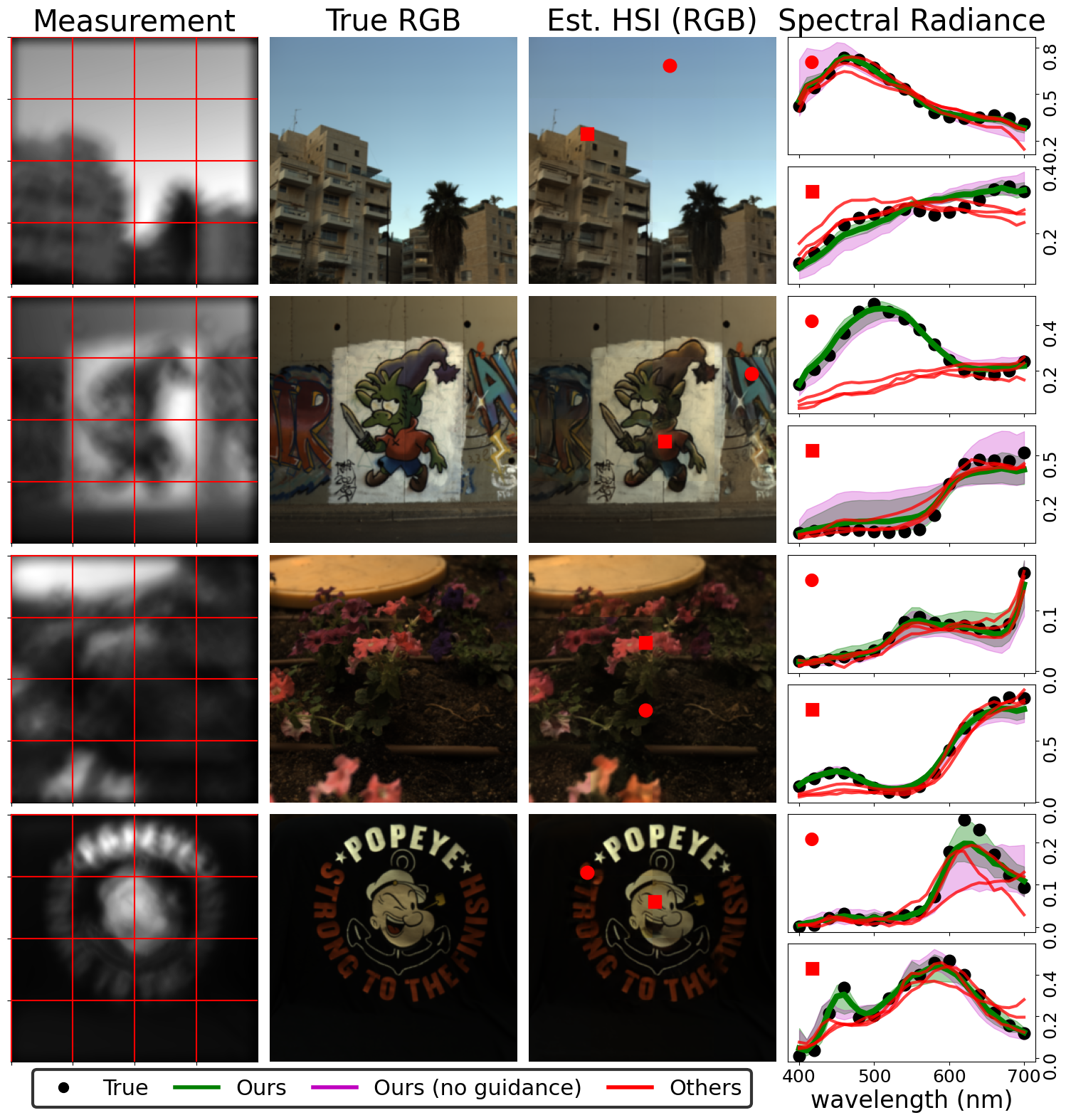}
            \caption{Left columns: grayscale measurement, ground truth RGB, and our estimated HSIs displayed in RGB. Right column: spectral profiles at two marked pixels per scene. Green: our mean reconstructed estimate with uncertainty (fill). Magenta: without guidance. Red: three next-best baselines. Black dots: ground truth.}
            \label{fig:g2h_main}
        \end{subfigure}
        \hfill
        \begin{subfigure}[t]{0.495\linewidth}
            \centering
            \includegraphics[width=\linewidth]{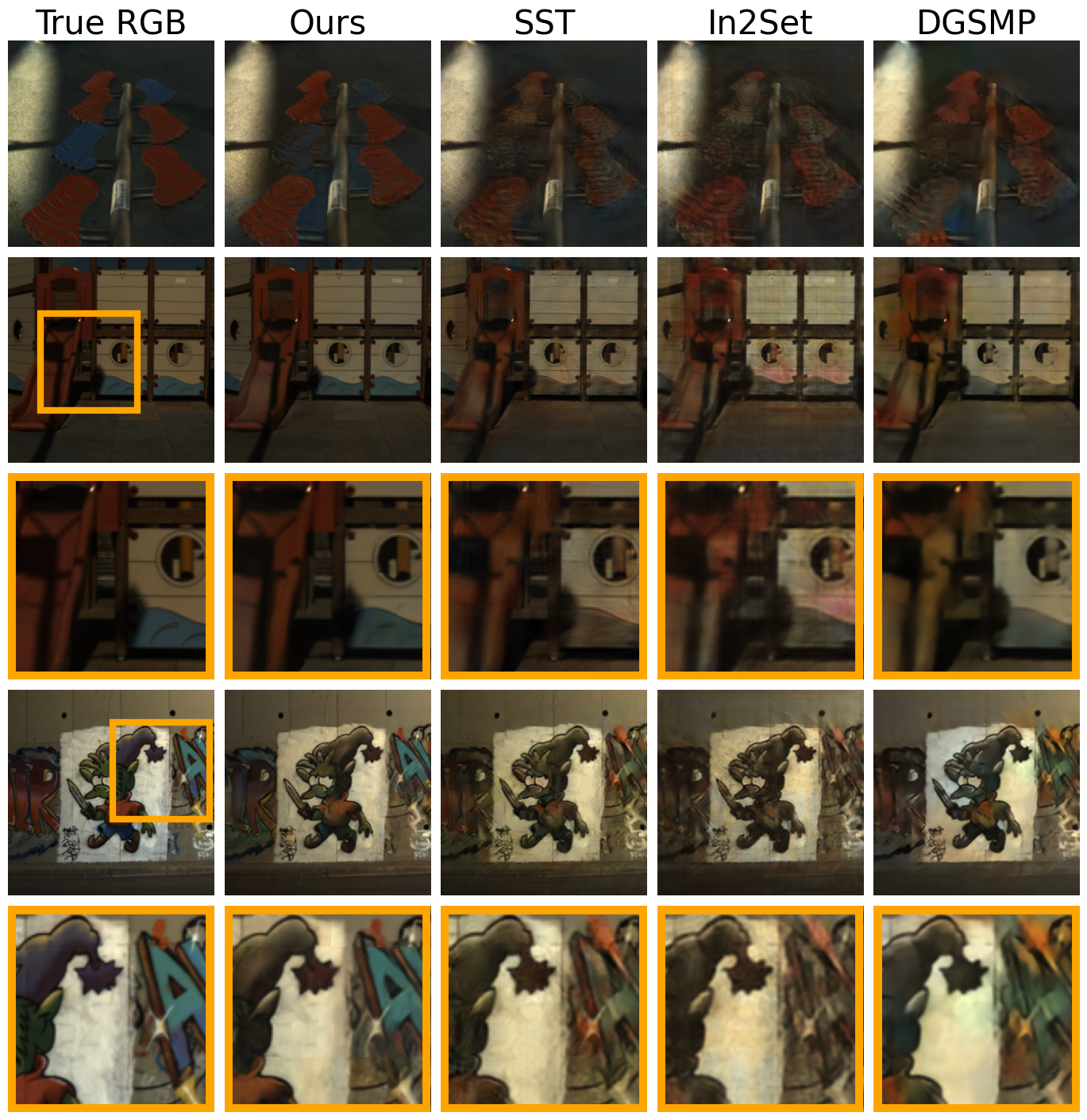}
            \caption{Comparison of reconstructions (projected to RGB) from our model and leading baselines for three test scenes. Zoomed-in regions (orange squares) are shown in the third and fifth rows.}
            \label{fig:g2h_comparison}
        \end{subfigure}
        \caption{Simulated grayscale-to-hyperspectral reconstructions on the ARAD1K test set.}
        \label{fig:qualitative_results}
    \end{figure}
    We compare our patch-based diffusion approach to eight state-of-the-art hyperspectral reconstruction models~\cite{Yuanhao2022, SPECAT, Wang_2024_CVPR, cai2022b, huang2021, hu2022, Meng2020, SST} that map fixed-resolution, full-field measurements to full-field HSIs. As no prior models exist for our single-optic filterless setting, these represent the closest available baselines from related but better-conditioned snapshot HSI tasks. We train all models from scratch on our rendered measurements following their original training procedures, making only minimal modifications where necessary, \eg replacing their forward/adjoint operators with our measurement function and changing the number of output channels. In Supp.~Sec.~S4.2, we repeat this comparison using RGB Bayer-filtered measurements, where our method also achieves the best performance.
    

    The filterless results are shown in \cref{tab:ARAD_Table} and visualized in \cref{fig:qualitative_results}, with additional visualizations in Supp.~Sec.~4.5. Our model achieves an average PSNR of $34.63$\,dB, surpassing the next best method by $2.86$\,dB. We also obtain a higher SSIM ($0.94$ vs.\ $0.90$) and a lower SAM ($0.11$ vs.\ $0.15$), reflecting more accurate per-pixel spectral predictions. For comparison, we also train our model's UNet backbone as a single-shot reconstruction network using a standard MSE regression loss on the full HSI instead of patches (``UNet-Regress'' in \cref{tab:ARAD_Table}). This variant under-performs all models, suggesting that our gains come not from increased backbone capacity but instead from our patch-based diffusion. These results also suggest that focusing model capacity on local optical cues in patches while enforcing global consistency through guidance is more effective than processing the entire measurement with a larger receptive field.  
    \begin{figure}[t]
        \centering
        \includegraphics[width=\textwidth]{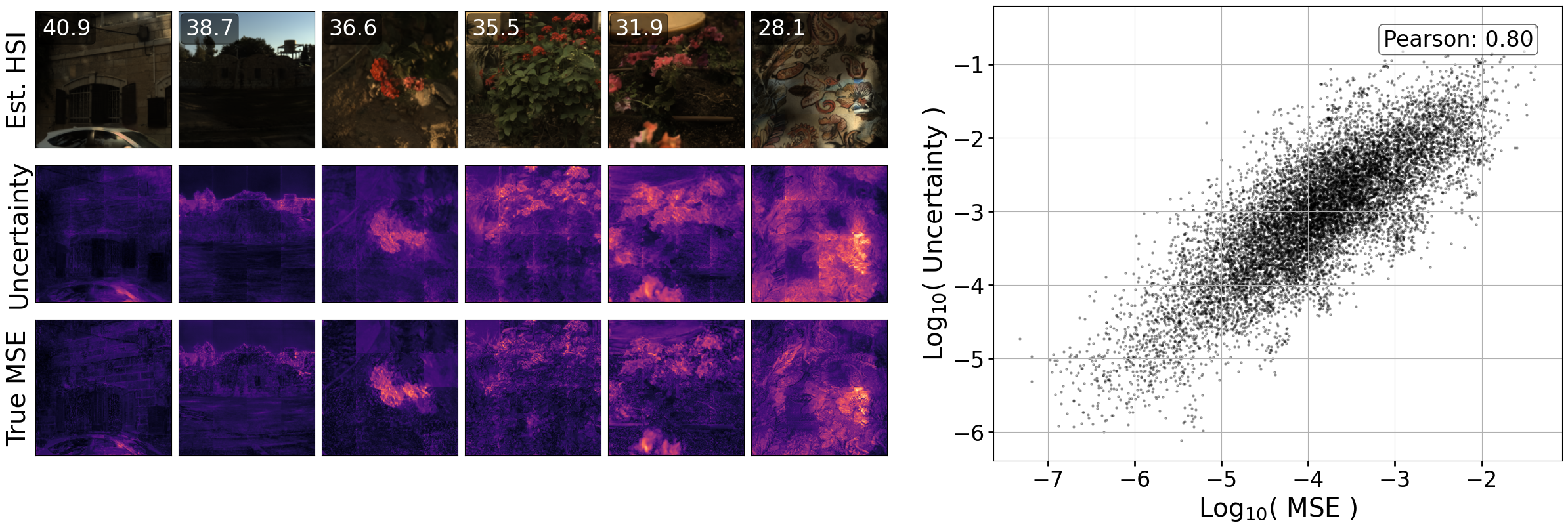}
        \caption{(Left) Estimated HSIs projected to RGB with per-pixel uncertainty maps and true MSE. Images are overlaid with PSNR and arranged from highest on the left to lowest on the right. (Right) Scatter plot of estimated uncertainty vs.\ true error from sampled test pixels.}
        \label{fig:sim_uncertainty}
    \end{figure}    
    \begin{figure}[t]
        \centering
        \includegraphics[width=\textwidth]{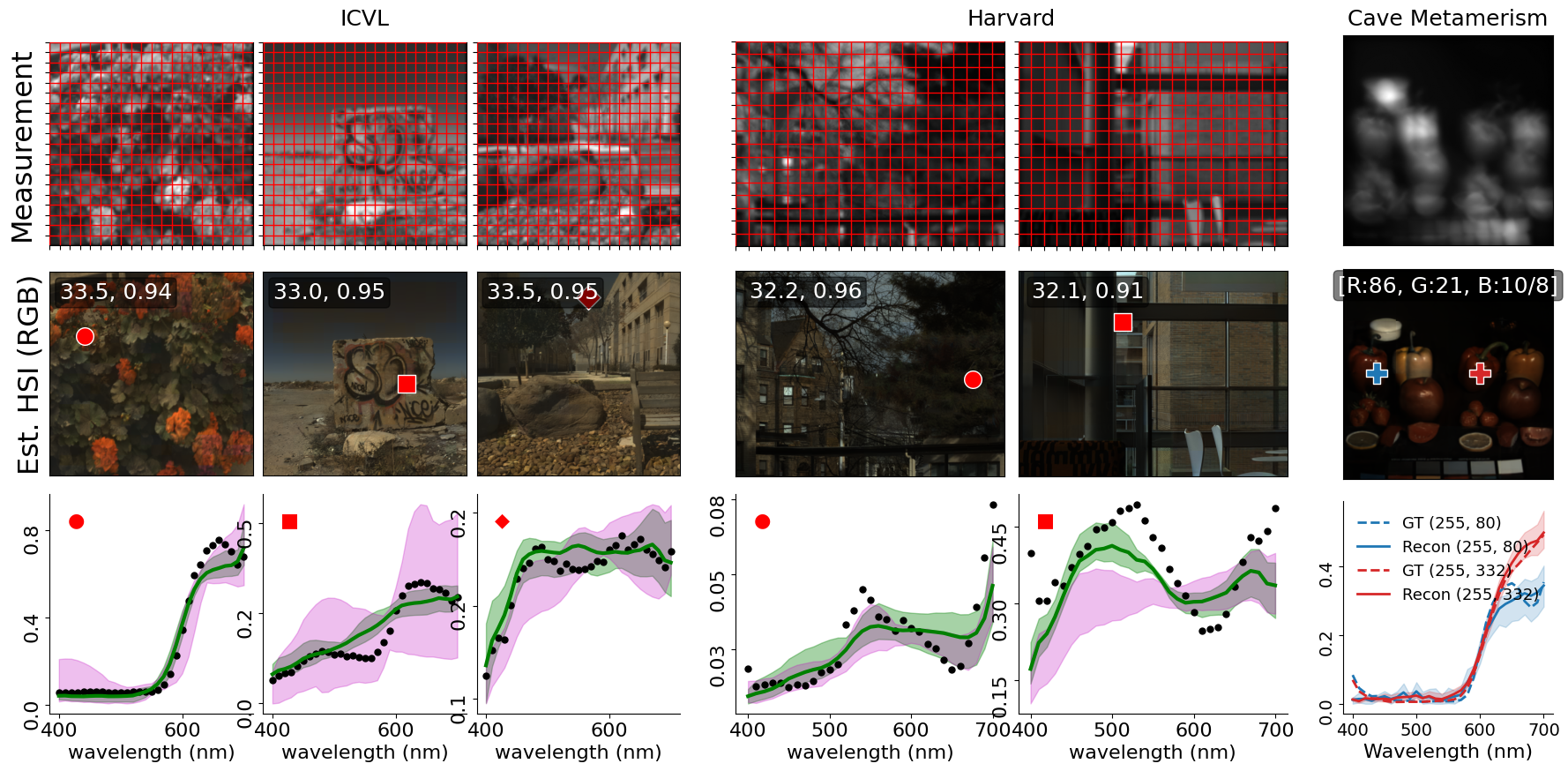}
        \caption{Cross-dataset reconstructions for ICVL (left), Harvard (middle) and CAVE (right, downsampled to $256\times256$ to reveal blurring). \textbf{Top:} measurement, with the patch grid in red for ICVL and Harvard. \textbf{Middle:} reconstructed HSI projected to RGB, with markers at the sampled pixels; PSNR/SSIM are overlaid for ICVL and Harvard, and the RGB values of the metameric pair for CAVE. \textbf{Bottom:} per-pixel spectral radiance at the sampled pixels, with fill showing the predicted uncertainty (cf. \cref{fig:sim_uncertainty}).}
        \label{fig:sim_crossdataset}
    \end{figure}
    
    In contrast to other models, our method also provides pixel-wise uncertainty estimates. These uncertainty maps, computed via \cref{eq:uncertainty}, are shown for several test scenes in \cref{fig:sim_uncertainty}. We find that the per-pixel uncertainty is strongly correlated with the mean squared error (MSE) between the predicted and ground-truth HSIs. Specifically, we compute a Pearson correlation coefficient of $0.80$ using 12.5K randomly sampled pixels from the 50 test images. This correlation remains strong when analyzed per-wavelength channel (Supp.~Sec.~S4.1). These results suggest that the uncertainty estimates may be useful in practice for identifying regions where the reconstruction is less reliable.

    \subsection{Cross-Dataset Generalization}
    We apply our diffusion model to reconstruct HSIs from other datasets, some at different spatial resolutions, without further finetuning. We train on ARAD1K at its native resolution of $512\times512$ pixels and then reconstruct measurements rendered from ICVL~\cite{Arad2016}~($1280\times1280$; 400 patches), Harvard~\cite{chakrabarti2011}~($1024\times1344$; 336 patches), and CAVE~\cite{Yasuma2010}~($512\times512$, 64 patches). \Cref{fig:sim_crossdataset} displays the reconstruction results for selected scenes, with additional visualizations in the supplement. Averaging over 10 test scenes, we obtain a mean PSNR/SSIM of 33.48/0.94 for ICVL and 32.37/0.92 for Harvard, indicating that our model generalizes well when the optical model is unchanged. We also observe wider uncertainty spreads in the predicted spectra (magenta and green fill in \cref{fig:sim_crossdataset} as compared to in \cref{fig:g2h_main}), reflecting the model's increased uncertainty under distribution shift. In the rightmost column of \cref{fig:sim_crossdataset}, we highlight our model's reconstruction when applied to an RGB-metameric scene from the CAVE dataset. Following \cite{fu2025limitations}, we reconstruct a scene containing real and fake fruits, where pairs of objects have spectral radiances that differs but RGB tristimulus values that are nearly identical. We find that our model can directly discriminate between metameric pixels without finetuning. We attribute this property to the fact that our reconstructions are driven by local optical cues (\cref{ssec:lens_comparison}) rather than learned object-level semantics.
    
    
\subsection{Run Time and Model Ablations} 
\label{ssec:model_ablations}
    \begin{figure}[t]
            \centering
            \begin{subfigure}[t]{0.495\linewidth}
                \centering
                \includegraphics[width=\linewidth]{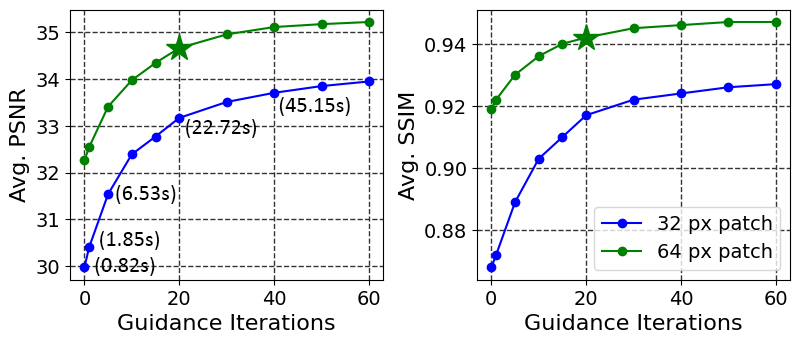}
                \caption{Reconstruction accuracy vs.\ number of guidance iterations for 32 and 64 pixel patches. Stars mark the default setting (20 iterations). Parenthetical values indicate runtime in seconds, which is approximately independent of patch size.}
                \label{fig:guidance_loops}
            \end{subfigure}
            \hfill
            \begin{subfigure}[t]{0.495\linewidth}
                \centering
                \includegraphics[width=\linewidth]{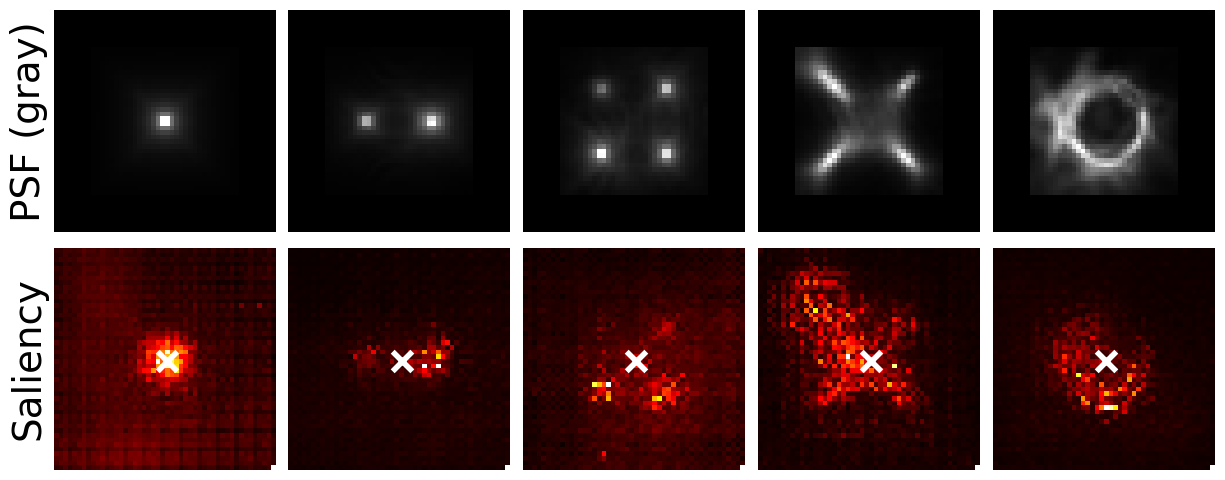}
                \caption{Saliency maps (bottom) for models trained with different PSFs (top, grayscale), showing which measurement pixels influence the spectral prediction at the probe location (white 'x'). The learned saliency patterns closely mirror the PSF structure.}
                \label{fig:InputSaliency}
            \end{subfigure}
            \caption{Guidance iteration analysis (left) and input saliency visualization (right).}
            \label{fig:guidance_and_saliency}
    \end{figure}
    We apply guidance by taking a gradient descent step on $\mathbf{x}_t$ (\cref{eq:guided1}) that is \emph{regularized} by the diffusion model, repeating it for several iterations before the next denoising step $\mathbf{x}_{t-1}$ (\cref{eq:guided2}). \Cref{fig:guidance_loops} shows that reconstruction accuracy improves logarithmically with the number of guidance iterations, with PSNR increasing from 32.32 (no guidance) to 35.22 at 60 iterations. Patches are processed in parallel, so complexity is dominated by the convolution in the measurement operator, which scales as $O(n \log n)$ for $n$ patches via FFTs. This guidance step is the main computational bottleneck; we use gradient checkpointing to reduce memory at the cost of longer run times~\cite{chen2016training}. Reconstruction requires approximately 22 seconds per scene for ARAD1K, but up to 52 minutes per scene for the higher-resolution Harvard and ICVL on an RTX 3090 GPU. Our method is thus suited to applications that require fast capture but not fast reconstruction. This cost can be reduced in several ways. We empirically find that guidance is more beneficial in later denoising steps, so the iteration count can be scheduled across steps rather than held fixed. For example, increasing the iteration count via a quadratic schedule over the DDIM steps reduces reconstruction time by $3\times$ on the ARAD1K test set without a loss in quality. Our method is also compatible with diffusion distillation techniques such as DMD~\cite{yin2024onestep}, which can reduce sampling to a few steps, offering a path from minutes to seconds per scene.    

    \Cref{tab:ablation} illustrates the effects of other design choices. In addition to toggling guidance, we evaluate smaller patch sizes, overlapping patches, and disabling patch rescaling ($c_{\text{lsq}}^p=1$ in \cref{eq:clsq}). In the latter case, we trained a separate diffusion model without patch normalization (\cref{ssec:Diffusion}). Interestingly, reducing the patch size to 32 pixels---equal to the PSF kernel width---only marginally reduces performance, even though it makes the problem substantially more ill-posed. This shows that the cross-patch synchronization introduced by guidance plays a critical role in mitigating patch-based ambiguity, since removing it causes a larger PSNR drop for 32-pixel patches (33.27 to 29.27) than for 64-pixel patches (34.67 to 32.16). We also tested patches smaller than the PSF kernel size (e.g., 16$\times$16 pixels), but these resulted in substantially worse reconstructions. Overlapping patches provide little benefit when guidance is active but become important otherwise, confirming that guidance corrects boundary artifacts. Finally, computing the patch scale factors $c_{\text{lsq}}^p$ during sampling, instead of training the network to predict an exact per-patch scale, yields substantial improvements.
   
\subsection{Comparison of PSF Designs and Model Interpretability} 
\label{ssec:lens_comparison}
    \Cref{tab:lenscomp} shows how different PSF designs affect the reconstruction quality, using the eight PSFs depicted in \cref{fig:metalens}. For each lens, we render grayscale measurements using the ARAD1K dataset and train a separate diffusion model for the same number of steps. Empirically, we find that reconstruction accuracy increases for lenses that induce stronger spatial-spectral mixing, but only to a certain extent. The T4 and R1 PSFs yield the best results, while the heavier mixing in the R2 and R3 designs from~\cite{Jeon2019} cause a decline in performance, likely due to excessive blurring that diminishes spatial detail. These findings underscore the importance of balancing spatial and spectral encoding, and they suggest that the PSFs best suited for filterless hyperspectral imaging differ from those designed for RGB sensor mosaics~\cite{Jeon2019}.
        
    To interpret what the patch-based diffusion models learn, we compute saliency maps via perturbation~\cite{zeiler2013, simonyan2014}. For a probe pixel at location $(r_x,r_y)$ in an output HSI patch, we define the saliency of each input measurement pixel $(i,j)$ as $S(i,j \mid r_x, r_y) = \mathbb{E}_p\left[ \sum_\lambda \lvert \partial \mathbf{x}_0^{p}(r_x, r_y, \lambda) / \partial \mathbf{y}^{p}(i,j) \rvert \right]$. We approximate this by systematically setting individual measurement pixels to zero, re-running the reconstruction for the patch (without guidance), and recording changes in the output spectrum at the probe location. We average the resulting saliency maps over 20 randomly sampled patches from the test set. As shown in \cref{fig:InputSaliency}, the saliency map for each lens model closely resembles the PSF kernel used to generate the measurements, even though these kernels are not explicitly provided to the network during training. This suggests that our models implicitly learn the structure of the physical image formation process, prioritizing the pixels in the patch that optically map to each output location.
\section{Experimental Prototype}
\label{sec:prototype}
    \subsection{Calibration and Setup}
\begin{figure}[t]
        \centering
        \includegraphics[width=1.0\linewidth]{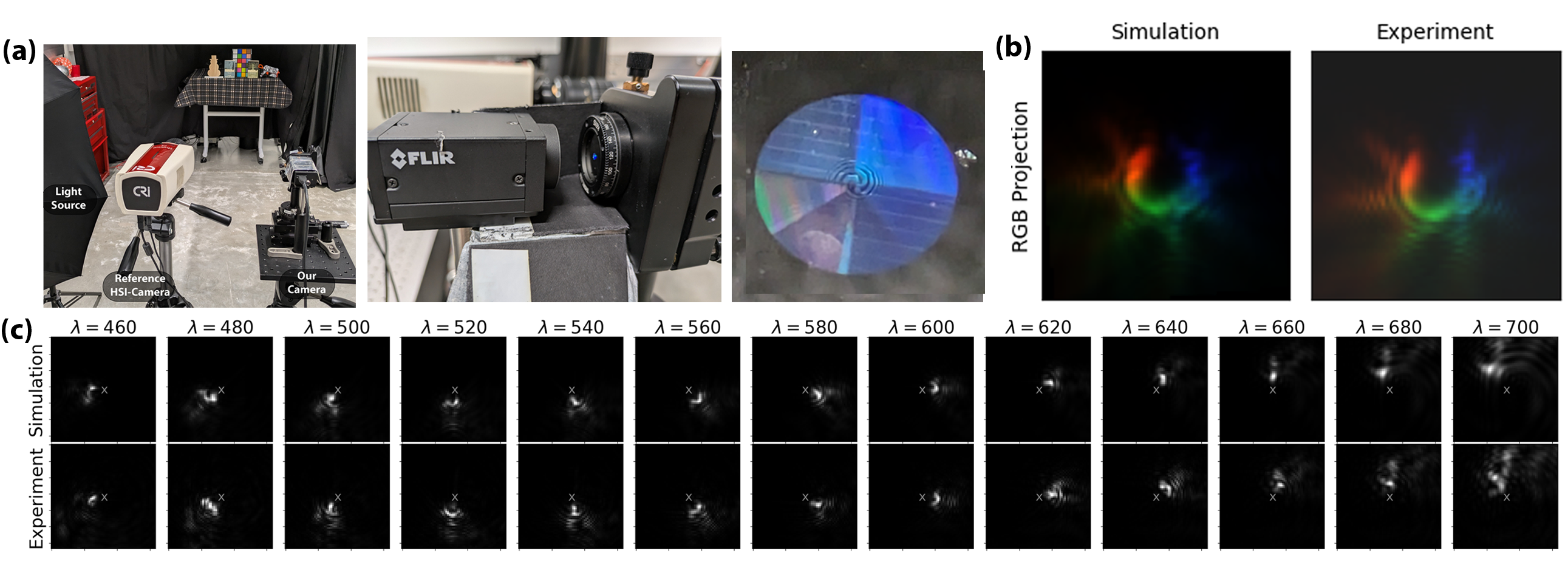}\\
        \caption{\textbf{(a)} Imaging setup: a reference hyperspectral camera and our metalens prototype capture tabletop scenes side by side. Closeups show the prototype camera and the fabricated metasurface. \textbf{(b,c)} Comparison of the experimentally measured and simulated R1 PSFs, shown as RGB projections (top right) and per-wavelength slices from 460 to 700~nm.}
        \label{fig:experiment_img}
    \end{figure}

    To validate our approach in a real imaging system, we construct the prototype shown in \cref{fig:experiment_img}(a). We fabricate a 3\,mm diameter metasurface composed of sub-wavelength $\mathrm{TiO}_2$ nanocylinders following the R1 PSF design (\cref{ssec:metalens}); see Supp.~Sec.~S2 for fabrication details. The metasurface is mounted 4\,cm in front of a monochrome CMOS sensor, with a 450--700\,nm band-pass filter at the aperture. \Cref{fig:experiment_img}(b,c) compares the simulated PSF to the experimentally measured PSF captured using a tunable laser source, confirming excellent spatial agreement. Approximately 96\% of the incident energy is concentrated within a $128\times128$ pixel ($832\times832$\,$\mu$m) region at the sensor, which we use as the PSF kernel. Compared to the $64\times64$ kernel used in simulation (\cref{sec:experiment}), this larger spatial support means that more light is scattered across patch boundaries when forming measurements, making the reconstruction more challenging. Separately, we calibrate the per-wavelength spectral scaling of the measured PSF using paired captures of planar calibration scenes; see Supp.~Sec.~S2.4 for details.
    
    We pre-train our model from scratch using the experimentally measured PSF, and then we finetune it to bridge the gap between simulation and the physical system, where stray light extends beyond the truncated PSF kernel, and where the sensor's spectral sensitivity and noise are not precisely known. To finetune, we need paired captures of real measurements and reference HSIs. For this, we place a commercial multi-shot non-compressive hyperspectral camera (Nuance FX, CRI Inc.) beside the prototype (\cref{fig:experiment_img}a) and capture 15 indoor tabletop scenes with both cameras, crudely registering the two views using per-scene homographies (full details in the supplement). Because these scenes contain objects at varying depths, the homography yields only approximate spatial alignment. We adapt our model to tolerate this misalignment by introducing random spatial offsets during patch extraction in pre-training. Specifically, when pairing a measurement patch with its reference HSI patch, we apply a small random shift so that residual misalignment is modeled as an additional source of per-patch ambiguity in the learned conditional distribution. The resulting jitter in per-patch predictions is corrected by guidance at inference, which enforces consistency with the full-field measurement after stitching the patch predictions. This allows us to finetune on these imperfectly aligned pairs and still bridge the remaining sim-to-real gaps in spectral calibration, stray light, and sensor noise.

    
\subsection{Training and Results}
\label{ssec:exp_results}
    \begin{figure}[ht!]
        \centering
        \includegraphics[width=0.96\linewidth]{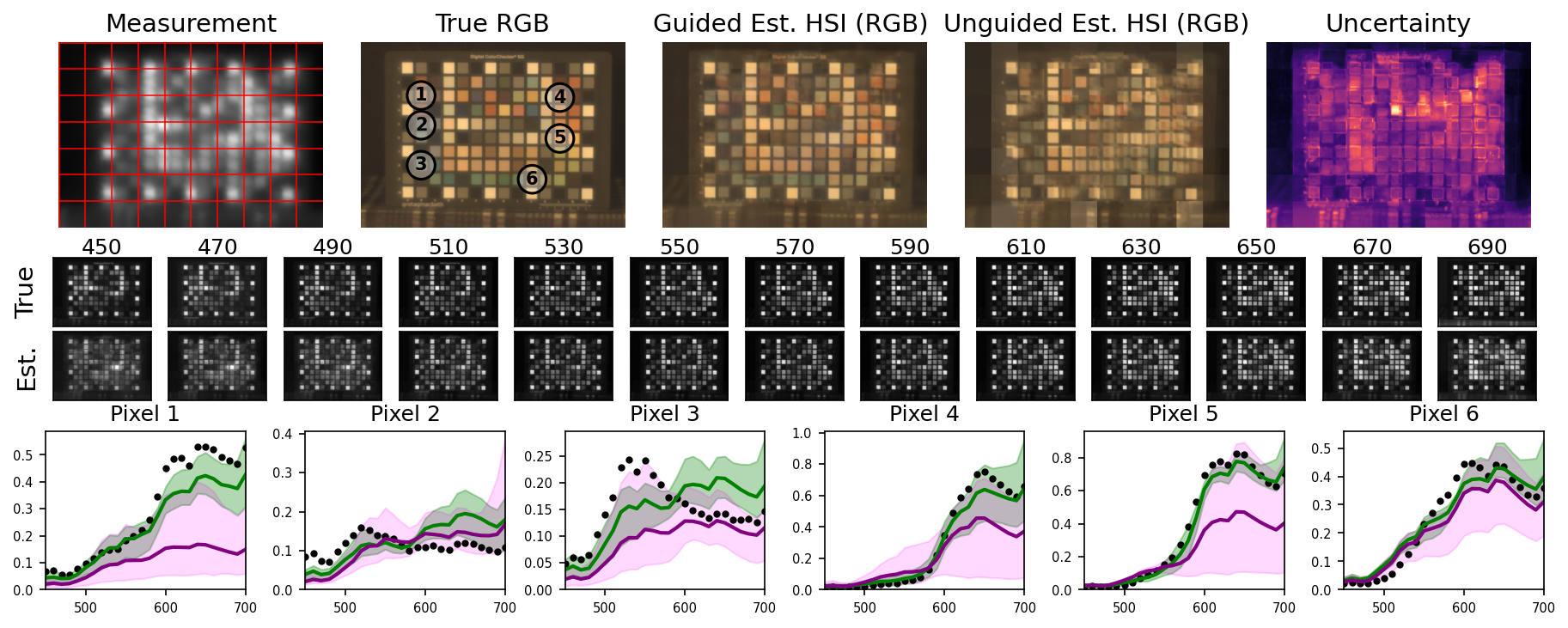}\\
        \includegraphics[width=0.96\linewidth]{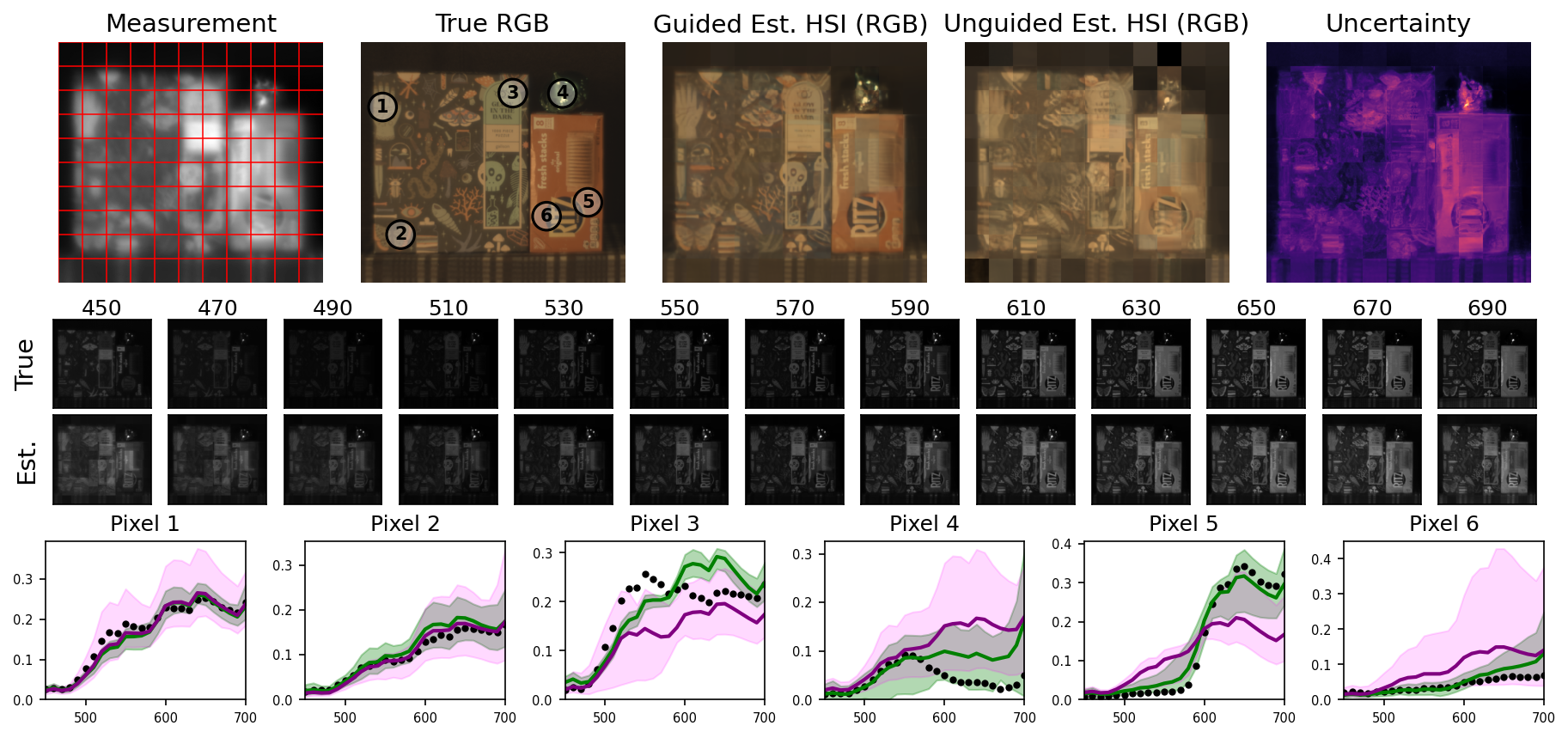}\\
        \includegraphics[width=0.96\linewidth]{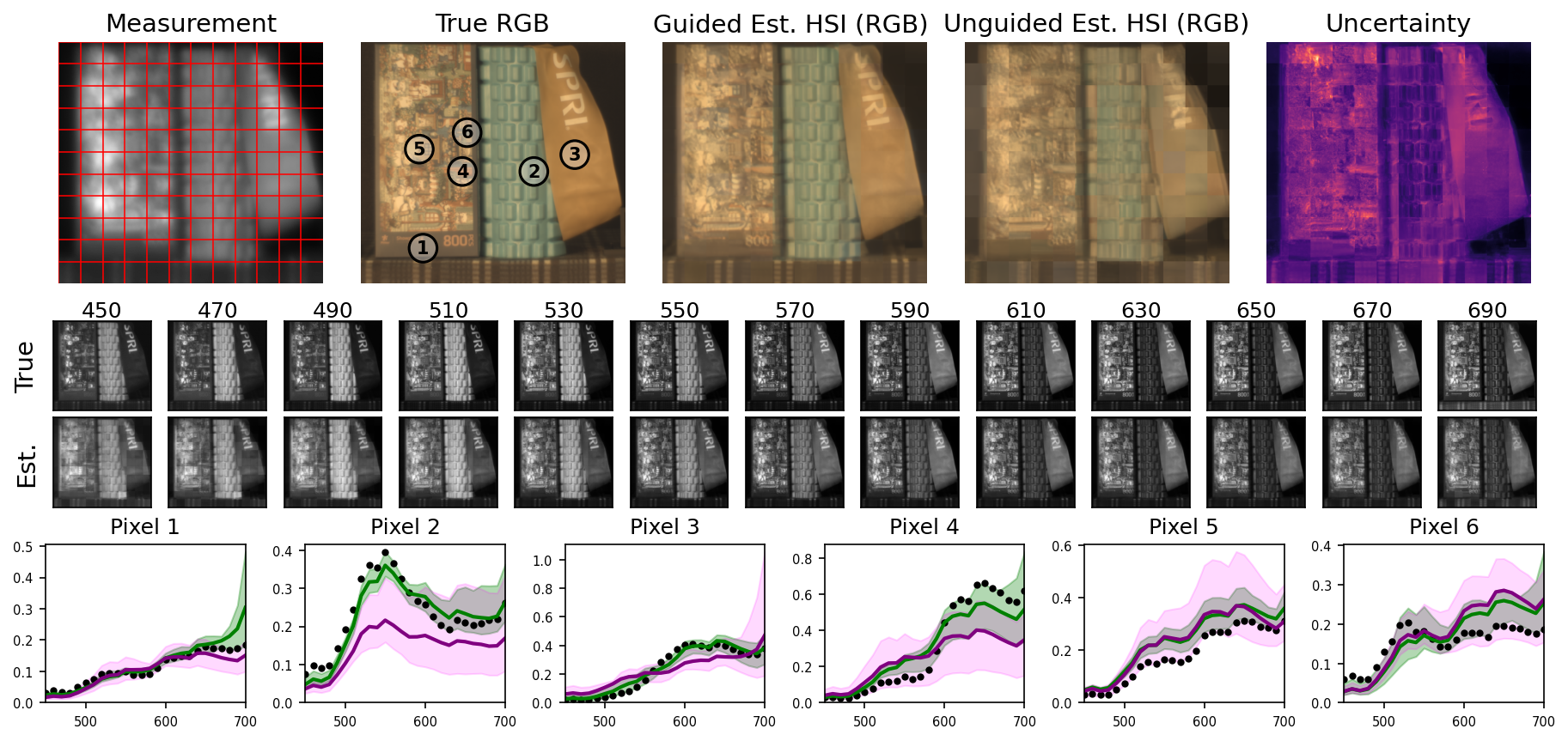}\\ 
        \caption{Experimental reconstructions for three test scenes. Top rows: grayscale measurement (with patch grid in red), ground truth RGB, guided and unguided HSI estimates (projected to RGB), and per-pixel uncertainty. Middle rows: per-wavelength slices of the ground truth and estimated HSI. Bottom rows: spectral radiance at six marked pixels using guidance (green) and unguided (magenta) with uncertainty (fill)}.
        \label{fig:exp_scenes}
    \end{figure}
    We pre-train the diffusion model on simulated HSI–measurement pairs rendered from the ARAD1K dataset using the scale-calibrated experimental PSF via \cref{eq:measurement}. We assume the PSF is shift-invariant, which we validate experimentally in Supp.~Sec.~S2.3. Training uses random $64\times64$ patch crops from the full-field pairs. To prepare the model for finetuning using the imperfectly aligned supervision pairs described above, we introduce random spatial offsets of up to 12 pixels when extracting these crops. Gaussian noise is also added to simulated measurements; see Supp.~Sec.~S3.4 for details.

    After pre-training, we freeze the model and finetune only the QKV attention matrices and the final projection head using rank-8 LoRA~\cite{hu2021lora, guo2025}, updating 166\,K parameters (0.22\% of the full model) with a learning rate of $1\times10^{-6}$ for 10 epochs. We use six of the 15 tabletop scenes for finetuning and evaluate on the remaining eight. The finetuning and test scenes contain no shared objects. For each test scene, we reconstruct 26 spectral channels spanning 450–700\,nm over a $640\times960$ pixel central crop (150 patches) of the full sensor measurement, corresponding to the region containing the scene objects. We use 10 guidance iterations to avoid over-guiding on noisy measurements; all other settings match \cref{sec:experiment}. Reconstruction requires approximately 13 minutes per scene. 
       
    \Cref{fig:exp_scenes} displays results for representative test scenes, with all scenes and full metrics in Supp.~Sec.~S4.3. Across the test set, our model achieves an average PSNR of 31.20\,dB, SAM of 0.157 ($8.9\degree$), and SSIM of 0.894. These results confirm experimentally that HSIs can be reconstructed solely from lens-induced chromatic aberration in a filterless snapshot measurement. 
    \begin{figure}[t]
        \centering
        \includegraphics[width=0.90\linewidth]{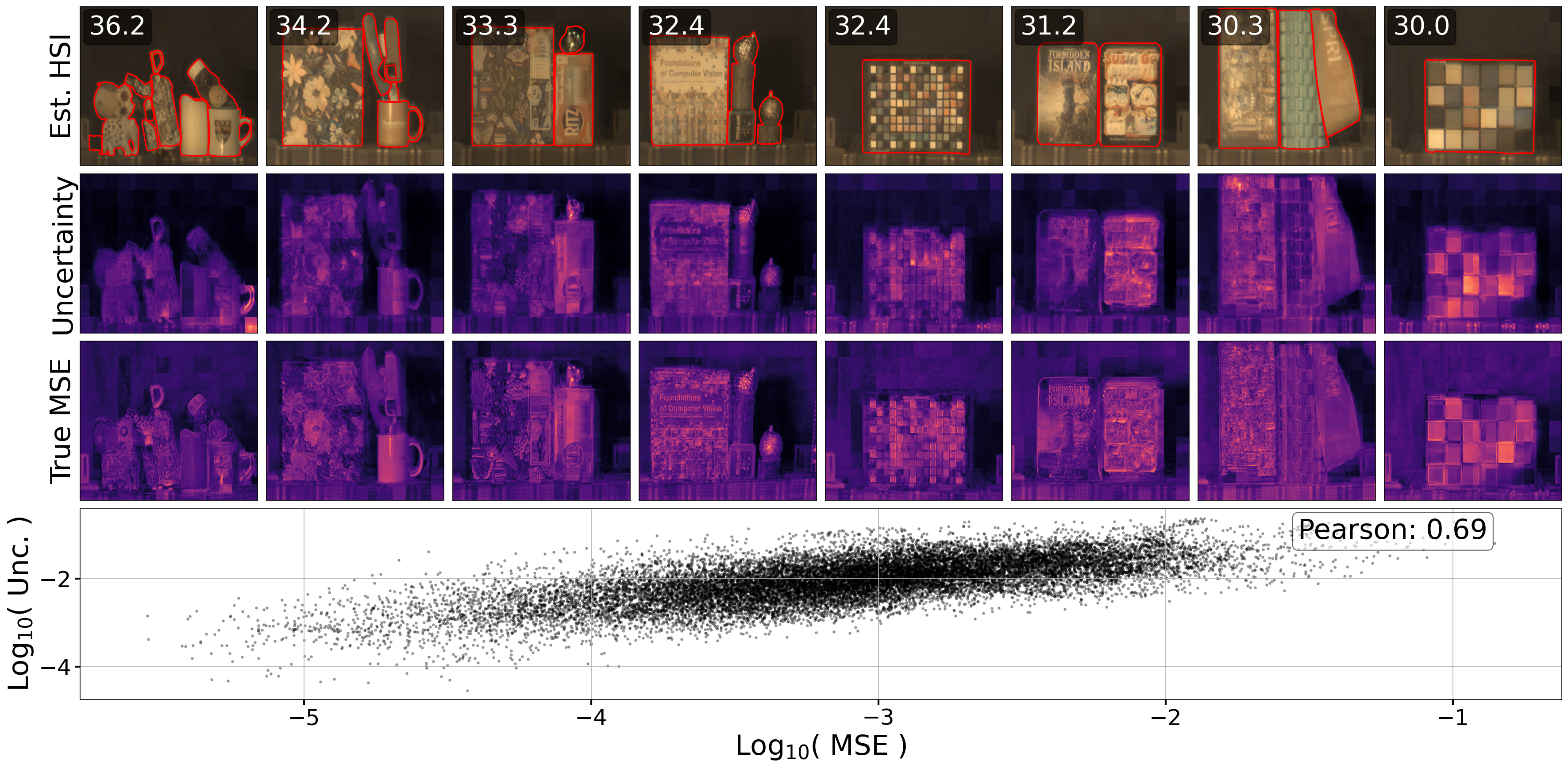}
        \caption{Uncertainty versus reconstruction error on experimental test scenes (cf. \cref{fig:sim_uncertainty}).  Top row: guided HSI estimates projected to RGB with PSNR overlaid.}
        \label{fig:exp_uncertainty}
    \end{figure}
    
    Without guidance, the predicted spectra at each pixel varies substantially when repeating the reconstruction using different initial noise seeds (magenta fill in \cref{fig:exp_scenes}). Straight edges appear jagged and textured regions exhibit ghosting artifacts, as each patch prediction carries a random spatial offset. When guidance is applied, both effects are corrected by enforcing consistency with the full-field measurement: spatial artifacts are resolved and the spread of per-pixel spectral predictions becomes significantly narrower (green fill), often enveloping the true spectral radiance curve. As shown in \cref{fig:exp_uncertainty}, the predicted uncertainty is strongly correlated with reconstruction error. We sample 20K pixels from regions containing objects (indicated by the red contours) and compute a Pearson correlation coefficient of 0.69 between uncertainty and MSE, confirming experimentally that per-pixel uncertainty identifies regions where reconstruction is less trustworthy.
    
    \section{Limitations}
    Several limitations stem from the patch-based formulation. First, like all methods that encode spectral information into spatial structure via PSF blurring, our approach requires scenes with texture at the scale of the PSF support. A spatially homogeneous region produces a near-uniform measurement regardless of its spectrum, making the mapping from measurement to HSI many-to-one and the spectrum unrecoverable. The influence of spatial feature size on reconstruction accuracy is probed in a controlled synthetic experiment, where we find that reducing the spatial frequency of a checkerboard pattern increases reconstruction error (Supp.~Sec.~S4.7). This is also visible in the two ColorChecker test scenes (\cref{fig:exp_uncertainty}) where larger uniform squares are reconstructed notably worse than smaller ones. Second, because patches are reconstructed independently and guidance couples them only at shared edges, global context cannot propagate across the image, leading to uneven reconstruction quality within a scene (e.g., the red flowers in \cref{fig:sim_crossdataset}). An encoder that conditions each patch on a full-resolution image embedding could address this.

\section{Conclusion}
\label{sec:conclusion}
We present one of the first demonstrations that hyperspectral images can be reconstructed solely from the local chromatic aberration in a single grayscale measurement that is captured through a flat-optic lens. Central to this is our integration of patch diffusion models with guidance based on the camera's point-spread function. By leveraging local diffusion while enforcing cross-patch consistency, this work provides a new approach for processing optically encoded measurements, as well as a new tradeoff between optical and computational complexity in snapshot hyperspectral imaging.
\section*{Acknowledgements}
    This work was supported in part by NSF cooperative agreement PHY2019786 (an NSF AI Institute, iaifi.org). 

\bibliographystyle{splncs04}
\bibliography{main}

\includepdf[pages=-]{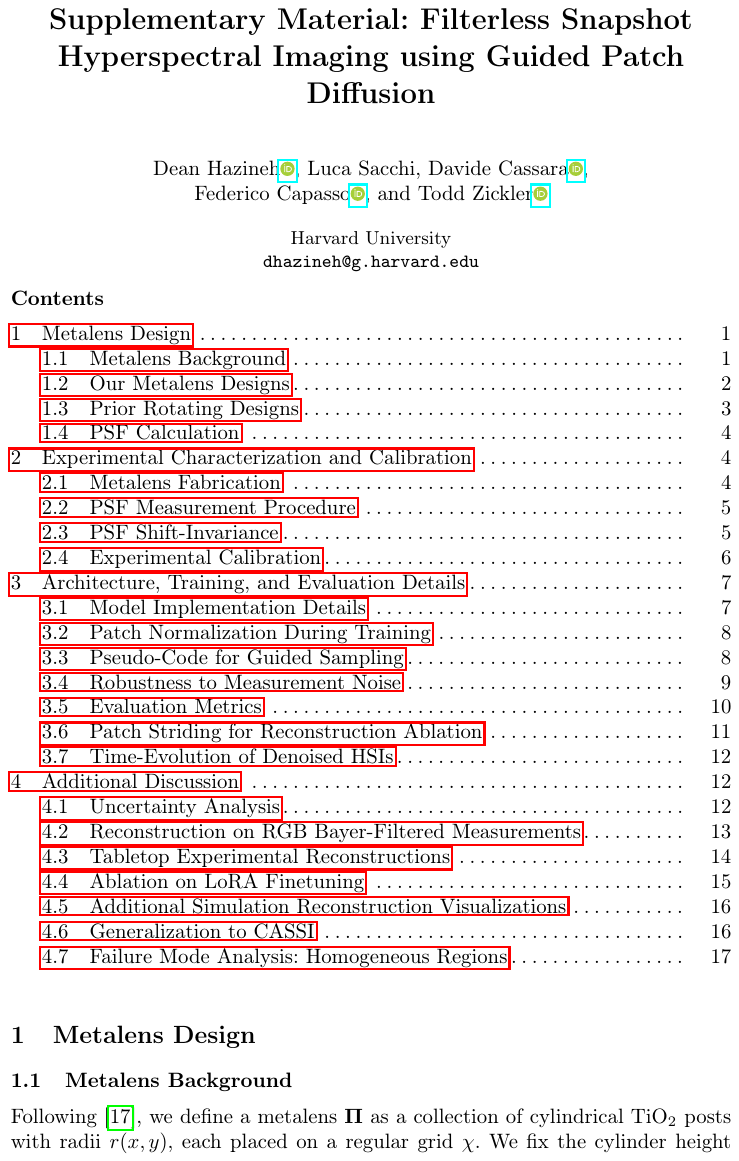}

    
        

\end{document}